\documentclass{article}

\usepackage{microtype}
\usepackage{graphicx}
\usepackage{amsmath}
\usepackage{subfigure}
\usepackage{booktabs} 
\usepackage{multirow}
\usepackage{hyperref}

\usepackage[accepted]{mlsys2020}

\mlsystitlerunning{A Predictive Autoscaler for Elastic Batch Jobs}

\begin{document}

\twocolumn[
\mlsystitle{A Predictive Autoscaler for Elastic Batch Jobs}



\mlsyssetsymbol{equal}{*}

\begin{mlsysauthorlist}
\mlsysauthor{Peng Gao}{heywhale}
\end{mlsysauthorlist}
\mlsyscorrespondingauthor{Peng Gao}{mike@heywhale.com}

\mlsysaffiliation{heywhale}{Heywhale, Shanghai, China}

\mlsyskeywords{autoscaling, cloud computing, time series prediction, machine learning}

\vskip 0.3in

\begin{abstract}
Large batch jobs such as Deep Learning, HPC and Spark require far more computational
resources and higher cost than conventional online service.
Like the processing of other time series data, these jobs possess a variety of characteristics such as trend, burst, and seasonality.
Cloud providers offer short-term instances to achieve scalability, stability, and cost-efficiency.
Given the time lag caused by joining into the cluster and initialization,
crowded workloads may lead to a violation in the scheduling system.
Based on the assumption that there are infinite resources and ideal placements available
for users to require in the cloud environment,
we propose a predictive autoscaler to provide an elastic interface for the
customers and overprovision instances based on the trained regression model. We contribute
to a method to embed heterogeneous resource requirements in continuous space into discrete resource buckets
and an autoscaler to do predictive expand plans on the time series of resource bucket counts.
Our experimental evaluation of the production resources usage data validates
the solution and the results show that the predictive autoscaler
relieves the burden of making scaling plans, avoids long launching time at lower cost and
outperforms other prediction methods with fine-tuned settings.

\end{abstract}
]




\section{Introduction}
\label{introduction}

In our production environment, we provide an elastic computing pool at scale,
where users can acquire batch jobs of any resource granularity.
However, these jobs are unlike to keep running all the time as online service.
With an aim to achieve cost-efficiency,
we dynamically scale up instances from the cloud provider and do the reverse when instances become idle.
The system is built on the Kubernetes\cite{burns2016borg}, a container orchestration
with cloud provider auto-scaling capability. Given that cloud providers have complex underlying
architectures, the time to launch an instance and do the initialization varies.
The proposed predictive autoscaler makes scaling plans and overprovisions instances to avoid time delay,
of which the objective is to give reasonable scaling plans for the number and type of instances.
There exist a series of instances of different specifications in the cloud provider's supply list.
Simple strategies may affect the performance of resources utilization by provisioning large instances that
leaves spare allocation space for small jobs whereas small instances lead to starvation of large jobs.
In this scenario, we use an embed method to summarize the utilization of our system resource and
adopt a first fit increasing bin packing algorithm to categorize the resource requirements of batch jobs, to find
the best suitable instances combination to launch.

Batch jobs SLA (Service Level Agreement) is resource-sensitive.
Job performance has a positive correlation with the number of resources assigned.
Typically a distributed deep learning job require machines with multiple instances with GPUs equipped.
Furthermore, batch jobs are like other time series data
described in \cite{hyndman2018forecasting} which is featured with trend,
seasonality, and peak and have cyclic bursty and on-and-off characteristics as described in \cite{9102411}.
To meet the elasticity needs of large batch jobs in production,
we use an Transformer-based neural network architecture
which is trained on the historical usage
of batch jobs to predict the future workload.
A cluster autoscaler is implemented by using this model to launch instances ahead instead of triggering scale-up on demand.

Our objective is to optimize the balance between
the job initialization latency and the resource cost. 
For example, providing a big number of large instances always match users' needs,
wheres providing instances with relative reasonable sizes can save the cost.
But small instances will not contribute to the scheduling if the average resource requirements keep large.
Thus we also find base instance type by comparing the scale it makes from the the smallest size to largest size.
We evaluate our optimized prediction model based on
the resources usage collected from our own production cluster, Alibaba Fuxi\cite{lu2017imbalance} and Microsoft Philly
\cite{jeon2019analysis} which are all the orchestration systems for batch jobs.
Most of the workloads of Fuxi are MapReduce jobs which are memory sensitive.
The Philly holds deep learning workloads which are sensitive to GPUs.
Our trace is a mix of Spark jobs and Deep Learning jobs.
It turns out that this method suits in the general requirements of resources sensitive elastic batch jobs and 
outperforms the classical time series prediction model as well as other neural networks.

The rest part of this paper is structured as follows.
Section II introduces the background information of running
elastic batch jobs on the cloud environment especially with Kubernetes.
Section III reviews and studies the work regarding predictive autoscaling on the cloud.
In Section IV the method to categorize resource requirement in continuous space,
the predictive models, and the autoscaler system architecture are presented.
Section V addresses the design and evaluation of the experiments outcome analysis, and
the comparison with the other predictive methods. Section VI reviews and concludes the work
in this paper.
\section{Background}
\label{background}

\subsection{Cloud Environment Capability}

The instance pricing models of cloud providers normally can be categorized into three types: on-demand instance,
reservation instance and preemptible instance.
On-demand instance can be charged by second, suitable for trial or short-term usage.
Reservation instance: is charged on a yearly or monthly basis with relatively low cost, but needs to be reserved for a long range of time,
applicable for well-planned and stable workload service.
Preemptable instance fails to support long-time running jobs,
which may be evicted by on-demand or reserved instances; however, cost can be saved, even 90\% cheaper than average instances.

To approach a prospective running environment where start latency time is shortened with a relatively low cost,
on-demand or preemptable instances are overprovisioned for initialization. Spark\cite{zaharia2010spark}, the typical batch job system, can tolerate executor failure and recover from
the failure stage. Distributed deep learning training frameworks like Horovod\cite{sergeev2018horovod} with AllReduce elasticity support
can downgrade the training scale when part of the nodes disrupt. In this case, the spot instance can be used. 
For other jobs that fail to be resilient to failures, we fall back to use on-demand instances.
But, all in all, temporary instances for short term jobs have many advantages in terms of the cost
compared with maximum static allocation.

The cloud environment possesses an unlimited capacity for computational resources.
Compared with static resources pool, there will be no problem related to the constrain for CPU/RAM/GPU requests limit issues. We assume the cloud providers can supply the desired instances as we require. Additionally distributed deep learning jobs have different topology requirements. For example, in AWS, users can specify the placement group of a instance which is a tag to mark the rack where the instance should be assigned to. This enables users to get access to the
resources without worrying about the detailed topology
of the underlying hardware.

\subsection{Features of Elastic Batch Jobs}

\begin{figure*}[t]
\centering
\includegraphics[width=6.75in,height=2in]{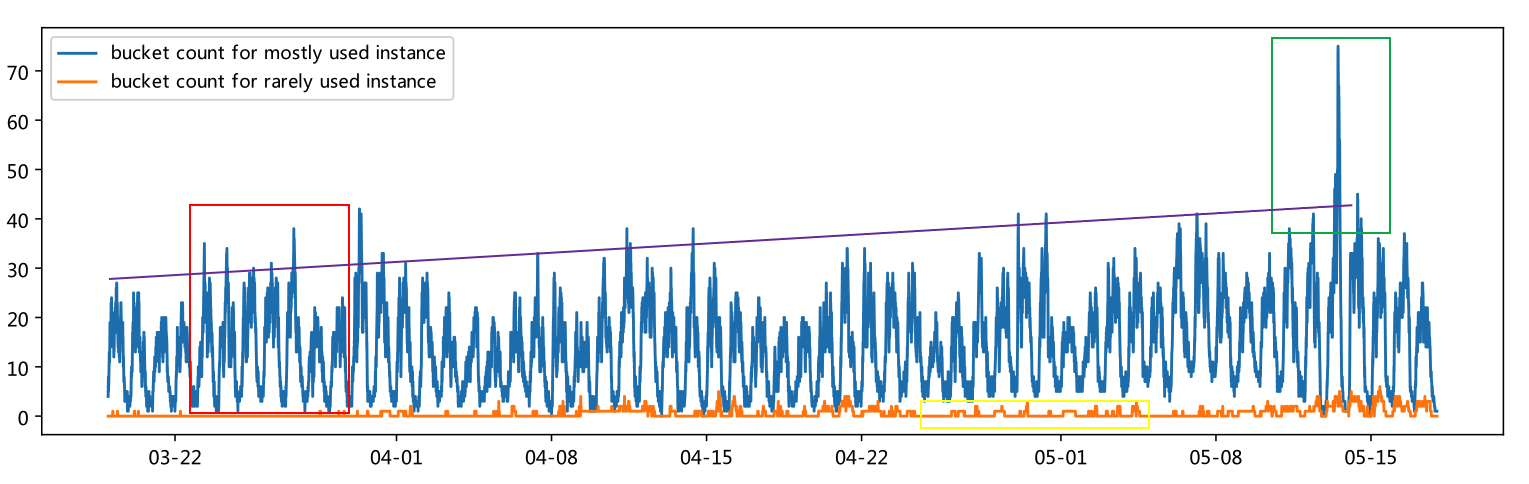}

\DeclareGraphicsExtensions.
\caption{A Resource bucket counts for both widely used and rarely used bucket counts. The pattern contains a week period, and reaches the peak at every middle of the week.}
\label{cpu_fig}
\end{figure*}

Elastic batch jobs require resources without knowing the underlying instances topology and placement. 
They assume the homogeneous capabilities of nodes, unawareness of the link latency during transferring load.
Besides, real-world batch jobs have a variety of workload patterns.
Figure \ref{cpu_fig} shows a widely used instances and a rarely used instances usage in our environment. The red rectangle marks the period of a week; The green rectangle marks the large burst of a temporary crowd usage; The yellow rectangle marks the on-and-off feature; The purple line marks a small trend over 3 months. Most batch jobs are triggered in workflows with a planned schedule,
which brings them the feature of seasonality; Other jobs are arised by a big event which leads to a crowd burst feature; Some batch jobs are trials or temporary use, which brings them the feature of on-and-off burst;
As business grow, basic usage gradually increases, which brings them the feature of trend. In other words, these features are more complex than general time series data.
This requires predictors to use an expressive model to deeply understand the patterns of workloads and enable autoscaler to give reasonable scaling decisions.

Different with online applications, offline batch workloads are resource-sensitive.
In our business model, the task is to ensure the resource allocation can be satisfied on time, while
users are responsible for application performance, who need to adjust the resource requirements
to tune the batch job performance, like the work \cite{or2020resource} which
adjust the scale of itself to find the best resources set for the job.
Meanwhile, resources requirements scale varies. Small shell scripts don't take up a lot of resources,
while large parallel computing jobs require distributed large size instances especially with GPUs equipped. With the development of deep learning, researchers have a strong desire to make quick experiments.
The increase of training work requires tens or hundreds of GPUs.
For example, Facebook used 256 GPUs to make a one hour training\cite{goyal2017accurate}, with a much higher price
than average work.
This brings us a challenge to establish a fine grain and cost-aware autoscaling strategy
where GPU requirements is the highest priority concern.

However researchers may have inconsistent result because of the miss placement. The problem comes from NUMA(Non-Uniform Memory Access) locality. Any memory directly connected to a CPU is considered being in the same NUMA node of a instance and can be accessed faster than other memories that not local to the CPU. This technology also extends to peripheral devices such as NICs or GPUs. The access speed depends on the how many interconnects must be passed through. All memory and peripheral devices on a NUMA system is divided into a set of NUMA nodes, with each node representing the local memory for a set of CPUs or devices. The same problem can occur for distributed workloads due to the topology of the links between servers. Workloads distributed on the instances in the same rack can utilize the high speed of the network connection, however cross-rack traffic goes through Ethernet. To gain higher throughput, 
distributed deep learning jobs tend to utilize the NVLink between the neighbour GPUs or the InfiniBand network connections between neighbour instances in the same rack. Cloud providers offer some options to organize instances in this way.

\subsection{Predictive Autoscaling}

\begin{figure}[t]
\centering
\includegraphics[width=3in,height=2in]{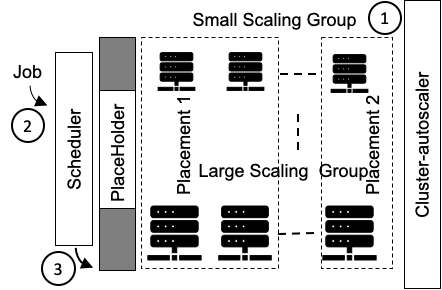}

\DeclareGraphicsExtensions.
\caption{Kubernetes Cluster-autoscaler overprovisioning}
\label{scaler}
\end{figure}

We used Kubernetes\cite{burns2016borg} as PaaS platform and Cluster-autoscaler\cite{kubernetesautoscaler}
as a component for automatic expansion. Different from application-level
scaling, cluster-wise scaling is mainly connected with instances,
which utilizes the API of the cloud provider to scale instance on demand to
join into the cluster. To be compatible with the cluster autoscaler and Kubernetes default scheduler which provide
preempt feature, we overprovision sets of placeholder applications which do nothing but sleep. Besides we assign them
with the lowest priority relative to other normal applications. Figure \ref{scaler} illustrates the scheme: \textcircled{\small 1} The cluster autoscaler triggers the scale-up of suitable underlying cloud autoscaling instance groups in different placement groups to add instances in the cluster when there are pending jobs (including placeholders) unable to run. There are strategies like random, least-waste, most-jobs. In order to fully utilize the resources, we use the least-waste option which is a first-fit bin packing algorithm. \textcircled{\small 2} When new jobs come, the scheduler will check for the idle resources as well as the preemptible placeholders as spare space for allocation. The placeholders will be preempted first instead of launching new instances on demand. \textcircled{\small 3} The placeholders have lowest priority, play a role to hold some idle instances and do nothing but sleep. The main focus of our predictor is to define the scale of placeholders. We need to make decision of certain numbers and types of placeholders in the recent time which should be smaller than the intilization time of new instances.

The predictive method follows the philosophy of dominant resource introduced in DRF\cite{ghodsi2011dominant}.
Dominant resource means the largest scale of a resource space. In this experiment, any required resource that occupies most of the instance resource
compared to other resources will be considered as dominant resource. For example, the dominant 
resource of a 1 core, 2GB RAM requirement for a 2 cores, 8GB RAM instance is CPU which takes 50\% of the instance
resource space. The first fit increasing bin packing algorithm is used to categorize the job's requirements in resource buckets
which are equivalent to cloud provider instances. We use increasing order because
small jobs won't trigger a scale-up of a large node but nodes of suitable size. If the method predicts lower usage
than current running jobs' requirements, it won't scale down the cluster but downgrade to an on-demand autoscaler,
since placeholders just can't reach negative replicas.
In the scale-up case, the predictor scales up the delta instances which is the subtraction to the current allocation snapshot.
\section{Related Work}

Many efforts have been devoted to solving problems regarding workload provisioning elasticity. The method
in \cite{roy2011efficient} is threshold-based scaling,
it makes regression on the performance metrics and scales up when
threshold reached. The threshold-based scaling is not dynamic enough to estimate heavily fluctuating workloads and doesn't estimate scale grain but the number of application replicas.

MLscale \cite{wajahat2019mlscale} presents a machine learning-based auto-scaling method, it relates
application-level monitored metrics with performance metrics and uses regression on monitored metrics
to predict scale.
These works focus on application autoscaling which is SLA-sensitive from the user's perspective,
while we focus on cluster wise autoscaling for batch jobs which are resource sensitive
from the provider's perspective.

RLPAS \cite{benifa2019rlpas} employs Reinforcement Learning which has an advantage that no training dataset is needed.
However it will take a longer time in the warm-up phase with a number of trials for stable performance.
In production, it is unacceptable. CloudInsight\cite{8968889} uses trace-driven simulation to generate data for autoscaling behavior learning.
After collecting a season of on-demand scaling data, we have enough confidence to
estimate future usage. In the beginning, a static ladder overprovisioning is used.
Upon the collection of a season of data, we switch to a predictive model.

Many statistical methods including seasonal exponential smoothing, ARIMA, and neural network have been
widely adopted to produce accurate results in \cite{roy2011efficient,mi2010online,yang2013workload},
Proactive auto-scalers in \cite{shariffdeen2016workload, 7847713} adopts an ensemble method to combine these predictors,
revealing that the neural network has
good performance for unknown workload patterns.

CloudInsight\cite{8968889} mentioned LSTM requires a massive amount of training dataset and computing resources.
With GPU equipped, the deep learning method can train fast, while there are no practical tools currently
available for training on GPU for statistical models, these models restricted with CPU slow down on large dataset
training.

Most of the existing autoscaling methods are threshold-based reactive
methods which scale the application resources based on single metric like CPU or Memory.
Our solution gives a concrete method to classify resources capacity and 
focuses on setting up an neural network
which is complex and expressive enough to learn general workloads patterns even unknown patterns.

\section{Method}

Our method consists of three parts: a method to embed continuous resources space into
discrete resource bucket counts, a neural network with Transformer architecture applied on
periodically collected resources usage data and a locality balancer for topology awareness autoscaling.
Our method predicts resources bucket counts and gives a corresponding scaling plan for every T minutes.
T is the average initialization time for the instances.

\subsection{Resource Embedding}

To classify resources usage, we define resource vector in multi-dimensional continuous space, in which each dimension represents a resource type. We define bucket vector as boundary to do the bin packing, the bucket size is set according to the corresponding instance type.

\begin{itemize}
   \item Resource Vector: a vector represents a job's resource requirements. Element
         can be continuous or discrete. For a job requiring 1 GPU, 3.5GB RAM, 1 CPU,
         the corresponding vector is $(1,3.5,1)$.
   \item Resource Bucket: a boundary bucket vector for allocation. It is the allocatable source
         equivalent to a certain instance type. Like an AWS m5.xlarge instance, it has 
         0 GPU, 8GB RAM, 2 CPU. So the bucket vector is $(0,8,2)$.
\end{itemize}

To apply first fit increasing bin packing on resources usage snapshot, we propose a comparison method,
which iterates over each element in the two vectors, if it meets a smaller element, then the
vector owning the smaller element is smaller and return, if it meets a tier, then continues. This procedure is
illustrated in the algorithm \ref{alg:comparison}.
The resource order in the vector represents
the cost factor of the resource. For the example of AWS, 
g4.xlarge with 4 cores, 16GB memory, 1 GPU is more expensive than m5.2xlarge with 8 cores and 32 GB memory.
If we want to trigger cheap instances for jobs requiring no GPU, leaving CPU and memory as spare resources,
the GPU element should be placed at first in the vector.

\begin{algorithm}[tb]
  \caption{Resource Vector Comparison}
  \label{alg:comparison}
\begin{algorithmic}
  \STATE $R = \langle r_i,\cdots,r_n\rangle $ \COMMENT{total resource space}
  \STATE {\bfseries Input:} resource vector $R_1$, $R_2$ in $R$
  \STATE $i=0$
  \REPEAT
  \STATE $r_1=R_{1i}$
  \STATE $r_2=R_{2i}$
  \IF{$r_1 < r_2$}
  \STATE $isLess = true$
  \ELSE
  \STATE i++
  \ENDIF
  \UNTIL{$isLess$ is $true$ or end of resource vector}
\end{algorithmic}
\end{algorithm}

Then we propose the embed algorithm \ref{alg:embed} to categorize resource allocations with greedy 
bin packing algorithm.
We compare each resource vector with resource boundary bucket, sum resource vector in the bucket which
the vector is less than, the count is the dominant resource division result. Table \ref{sorted_resource_vector} is an example in which there 4 sorted resource vectors and
two buckets $(0,2,1)$, $(1,4,2)$, the bucket counts sould be
calculated as table \ref{bucket_counts} illustrates.
In practice, we will round up all float counts to integers.
\begin{algorithm}[t]
  \caption{Embed Resource Requirements}
  \label{alg:embed}
\begin{algorithmic}
  \STATE {\bfseries Input:} sorted resource vectors $R$, sorted buckets $B$
  \STATE {\bfseries Output:} bucket counts $c$, length of $c$ = length of $B$
  \STATE $i$ = $0$
  \STATE $j$ = $0$
  \FOR{$i$ $<$ length of $B$}
  \FOR{$j$ $<$ length of $R$}
  \IF{$R_j < B_i$}
  \STATE $C_i$ += $R_j$
  \STATE $j$++
  \ELSE
  \STATE break
  \ENDIF
  \ENDFOR
  \STATE $i$++
  \ENDFOR
  \STATE $c_i$ = max $ C_{ij} / B_{ij} $ for $ j \in  R, i \in C$
\end{algorithmic}
\end{algorithm}

\begin{table}[t]
\caption{Sorted Resource Vectors.}
\label{sorted_resource_vector}
\vskip 0.15in
\begin{center}
\begin{small}
\begin{sc}
\begin{tabular}{ccc}
\toprule
GPU &  CPU & Memory \\
\midrule
  0 & 1 & 1GB\\
  0 & 1 & 2GB\\
  1 & 2 & 4GB\\
\bottomrule
\end{tabular}
\end{sc}
\end{small}
\end{center}
\vskip -0.1in
\end{table}

\begin{table}[t]
\caption{Bucket Counts}
\label{bucket_counts}
\vskip 0.15in
\begin{center}
\begin{small}
\begin{sc}
\begin{tabular}{cccc}
\toprule
GPU & CPU & Memory & Count \\
\midrule
  0 & 1 & 2GB & 1.5\\
  1 & 2 & 4GB & 1\\
\bottomrule
\end{tabular}
\end{sc}
\end{small}
\end{center}
\vskip -0.1in
\end{table}

After summarizing resource usage snapshot, we feed these bucket counts as time series input data
to our prediction model. The predictor uses the result to decide the strategy of autoscaling plan, and create or 
resize the placeholder replica sets to trigger the underlying autoscaling scheme of the cluster autoscaler.

\subsection{Workloads Prediction}

\begin{figure}[t]
\centering
\includegraphics[width=3in,height=2in]{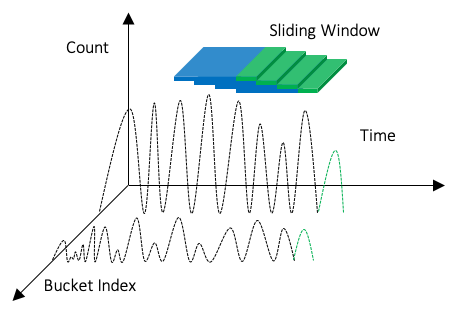}

\DeclareGraphicsExtensions.
\caption{Multi-dimensional one step ahead forecast}
\label{sliding_window}
\end{figure}

We use a sliding window to construct our train and test data. For one step ahead forecast, we use a fix length window to slide over our time series data as illustrated in Figure \ref{sliding_window}. The test data is the window next to the train window. Each element of our training data is the bucket counts with time index. We feed these data to the predictor and use the result to scale our delta instances.

Statistical models are more configurable than the neural network trained model, they do not need the data to be splitted into windows. However, information across similar time series cannot be shared since each time series is fitted individually. Further, they require detailed analysis for the parameters selection, and contain fewer parameters than the neural network,
which means fewer parameters can not highly reflect
the complex workload patterns. There are unknown and unstable changes which do not follow
the general features of time series data. 
The neural network contains a considerable amount of parameters
which are large enough to learn the patterns of workloads in the context of a large amount of datasets. The most popular methods of statistical models are exponential smoothing and ARIMA\cite{gardner1985exponential}. 

For neural network, the LSTM-based neural network is 
also widely applied in the time series forecasting.
It resolves the naive RNN network parameters vanishing problem. However, its performance degrades with long dependencies because it cannot adequately encode a long sequence into the intermediate vector. In such cases, how to model long-term dependencies becomes the critical step in achieving promising performances.

\begin{figure}[t]
  \centering
  \includegraphics[width=3in,height=2in]{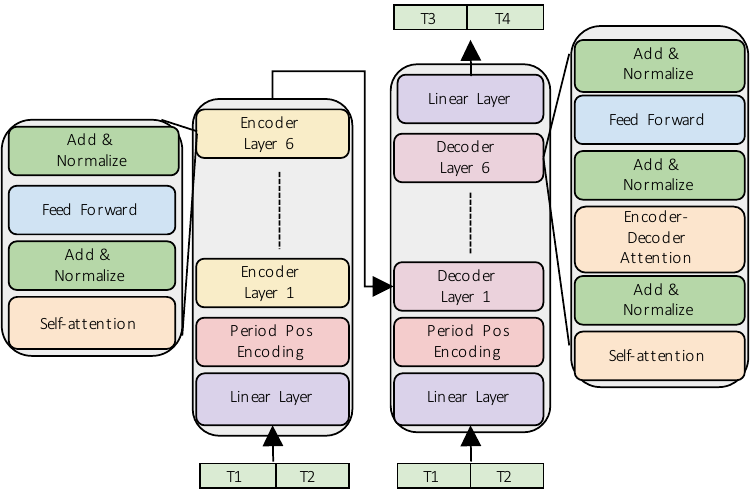}  
  \DeclareGraphicsExtensions.
  \caption{Transformer-based network architecture. The input linear layer flatten the input data to the $d_{model}$ dimension. The positional encoding layer is a periodic function by setting the season as period.}
  \label{transformer}
\end{figure}

We set up a Transformer-based neural network introduced in the original paper\cite{vaswani2017attention} which overcome the problems mentioned above, but modify the components to be compatible with time series data. The neural network architecture is illustrated in Figure \ref{transformer}.
It consists of input layer, encoder layer, decoder layer and output layer. The input layer is a linear layer to flatten the original time series vector to a $d_{model}$ dimensional vector which is then applied in multihead attention mechanism. The encoder layer consists of 6 identical encoder layers: a self-attention sub-layer and a linear layer. Both of them are followed by a normalization layer. We replace the original positional encoding layer with a periodic function where pos is the position in the window and n is the level of seasonality. 

\[ PE_{(pos)} = \sum_{i=1}^{n} sin(pos*2\pi/period_{i}) \]

The periods depend on the season of the data. For 5 minute data points, a period of one weak is 2016 and a period of one day is 288. The positional encoding vector is then used to encode the seasonal sequential information by adding to the input vector.

The decoder is also composed of the input layer, six identical decoder layers, and an output layer. The decoder input
begins with the last data point of the encoder input. The
input layer maps the decoder input to a $d_{model}$ dimensional
vector. In addition to the two sub-layers in each encoder
layer, the decoder inserts a third sub-layer to apply self-attention mechanisms over the encoder output. Finally, there
is an output layer that maps the output of last decoder layer
to the target time sequence. We employ look-ahead masking
and one-position offset between the decoder input and target output in the decoder to ensure that prediction of a time
series data point will only depend on previous data points.

For optimizer, We used the Adam optimizer with 
$\beta_{1}=0.9, \beta_{2}=0.98, \epsilon=10^{-9}$.
A custom
learning rate with the following schedule used:
\begin{multline*}
lrate=d_{model}^{0.5} \times min(step\_num^{0.5}, \\
step\_num \times warmup\_steps^{-0.5})
\end{multline*}
Where $warmup\_steps = 5000$. We set up a dropout of 0.2 for regularization.

\subsection{Topology Awareness Balance}

The GPU would make resource allocation decisions independent of each other in the Kubernetes. 
This could result in undesirable allocations on multi-socket systems, causing degraded performance on latency critical applications.
In the existing Kubernetes cluster scheduling algorithm, when the GPU needs to communicate with the CPU core, it will randomly select an idle CPU core to communicate. 
In order to reduce the unnecessary communication cost, the TopologyManager\cite{topologymanager} introduced in Kubernetes 1.18 bind GPU to the nearest n CPU cores.
The communication overhead will be reduced, since it will try best effort to schedule jobs aligned to the NUMA node in an instance. 

For intra-node alignment, we use a simple strategy to balance the jobs among the placement groups.
We define virtual instance which is a set of instances with the same size and in the same placement groups and zones.
We balance the overprovisioned virtual instances in the placement groups in a round-robin fashion to reduce allocation fragmentation and communication latency.

\begin{figure}[t]
  \centering
  \includegraphics[width=3in,height=2in]{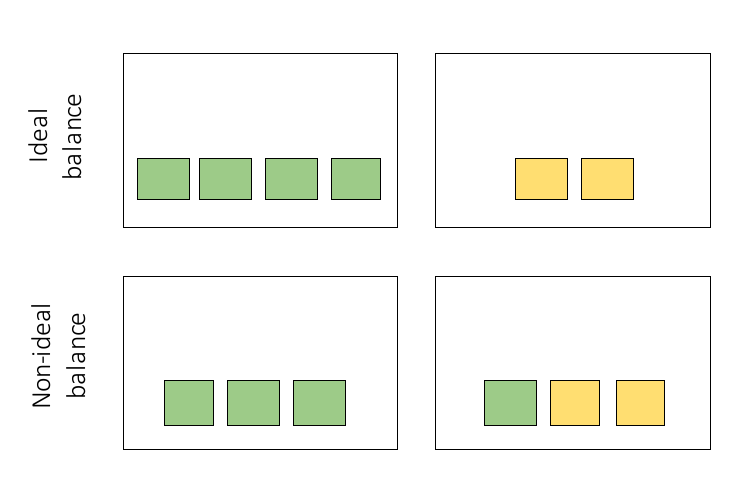}

  \DeclareGraphicsExtensions.
  \caption{Autoscaling Balance. The miss placement will degrade the performance of the distributed training jobs, and the ideal placement tends to balance the virtual instance in the same placement group.}
  \label{balance}
\end{figure}

If the number of GPUs requirements of jobs is larger than the number of GPUs of the largest GPU instance, we will trigger multiple instances scale-up.
In order to meet the locality requirement, we balance the instance over the placement groups, which then utilize the higher communication connection among the placement.
Figure \ref{balance} illustrates the situation: if there are 2 scale-up plan where a 4 instances and 2 instances in two virtual instances are going to be scaled up,
the ideal placement is that instances in the same virtual instance should be place in the same placement group instead of balancing each instances separately.

\subsection{Implementation}

\begin{figure}[t]
  \centering
  \includegraphics[width=3in,height=2in]{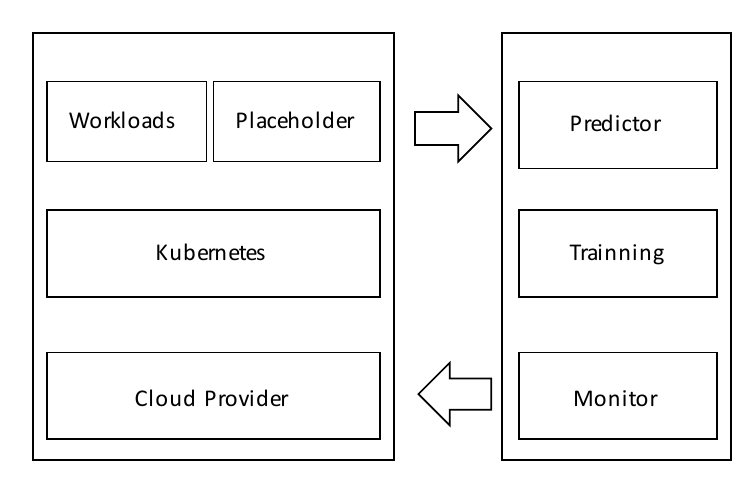}
  \DeclareGraphicsExtensions.
  \caption{Overall predictive autoscaler system architecture.}
  \label{sys_arch}
\end{figure}

The overall system architecture is illustrated in figure \ref{sys_arch}, it is based on
the Kubernetes and Cluster-autoscaler. For predictive overprovisioning,
we set up placeholder replica sets that equivalently occupy the 
specified instances. Our predictive autoscaling will not scale down below the current resource
requirements. Hence we subtract our prediction to current resources snapshot, only to scale up the placeholder
if the delta is larger than zero. Placeholders will work on the initialization work including image loading and
data files downloading, where time is saved if real workload comes in. The resource monitor collects the data related to workload resource utilization of the Kubernetes cluster,
and automatically queries auto scaling group from the cloud provider to set up bucket boundaries. 
The resource monitor sinks the collected usage to persistent storage, and a periodically triggered training job will input new data to refresh the regression model.

The prophet component queries the trained models and create or resize the replica sets based on the
predicted result. But at the initial stage, the prophet only uses a ladder scaling policy because of the unavailability of data. After being configured by the prophet, the placeholder replicas set
will scale which will then trigger cluster autoscaler to scale up instances from the cloud provider.

To pick a series of reasonable instances, we remove the smallest instance type from the series until the smallest instance stops contributing to cutting the scale. For example. In Figure \ref{res_util}, the base instance of Philly cluster is g4.2xlarge, because setting g4.4xlarge as the base instance will increase the scale. 

\section{Experiments}
\subsection{Dataset}
The overall dataset specification is illustrated in Table \ref{training_datasets}.
We evaluate our results on the real system of our own production and two other clusters with the assumption
that workloads are running on the cloud autoscaling environment instead of on-premise static environment.
We use the series of M5 and the series of G4 from the AWS to construct our bucket boundary.
M5 instances offer a general capability of compute, memory and networking, suitable for Spark and MapReduce jobs which are memory and CPU sensitive.
G4 instances deliver the industry’s most cost-effective and versatile GPU instance for deploying deep learning workloads\cite{awsg4instances}.
We aggregate the time series data by using a time tick of 5 minutes,
because empirically it generally costs no more than 5 minutes to make the instance launching and initialization on the AWS cloud.
The time tick can be adjusted with accordance to the launching delay assurance of other cloud providers.

\begin{figure*}[t]
  \centering
  \includegraphics[width=6.75in,height=2in]{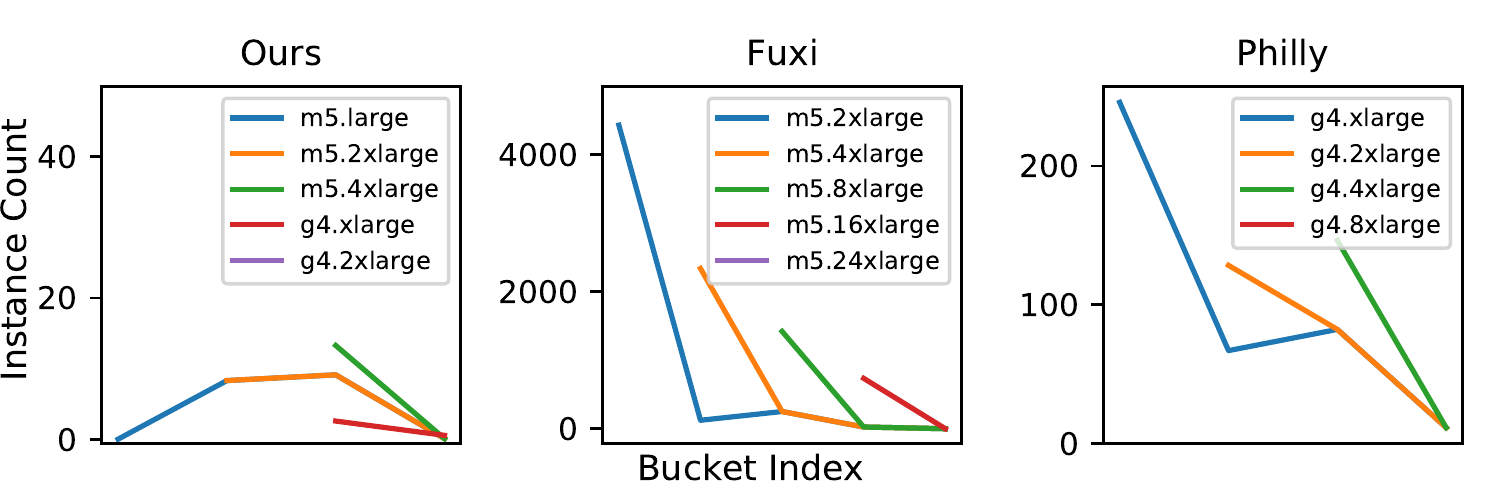}
  \DeclareGraphicsExtensions.
  \caption{Resource scale. Ours will utilize all the instances. Philly stops contributing to the scale at g4.4xlarge. Fuxi only uses m5.24xlarge since no smaller instances reduce the scale}
  \label{res_util}
\end{figure*}

\subsubsection{Fuxi Cluster}

Fuxi\cite{zhang2014fuxi}, the resource management and job
scheduling system that is capable of handling the kind of
workload at Alibaba where hundreds of terabytes of data
are generated and analyzed everyday to help optimize the
company’s business operations and user experiences. The trace has a time range of a weak. 
The original machines of the cluster are all the same size: normalized 100 RAM and 96 cores. 
Since we do not know the exact memory specification,
we assume the memory is the same size as the m5.24xlarge whose number of cores is also 96. 

Illustrated in Figure \ref{res_util}, we compute the scale made by m5.2xlarge, m5.4xlarge, m5.8xlarge, m5.16xlarge and m5.24xlarge by removing the smallest instance step by step.
Each line marks the scale the smallest instance makes, and it turns out that only m5.24xlarge is the reasonable instance type because other instances do not help to reducing the scale.
The time range of the cluster trace is one week, we use the beginning 6 days as our train set and the last day as test set.
It has a clear seasonality of one day, thus we set the period of the model as 288.

\subsubsection{Philly Cluster}

Philly\cite{234916} is deployed on large GPU clusters shared across
many groups in the company. 
The cluster has grown significantly over time, both in terms of the number of machines
machine and the number of GPUs per
machine. It also has high-speed network connectivity among
servers and GPUs in the cluster. To speed up distributed training where workers need to exchange model
updates promptly for every iteration, it requires jobs tend to run on the machines within the same rack
connected via 100-Gbps RDMA (InfiniBand) network, instead of letting cross-rack traffic go through Ethernet. This trace spans across two months and uses around
100,000 jobs run by hundreds of users.
The Philly cluster also has a large scale as the Fuxi cluster. We also apply our strategy to assume the workloads running on the cloud.
The cluster holds different deep learning training jobs. Some of them are distributed, and we use virtual instance introduced in Section 4.2 to categorize the resources.

\subsubsection{Our cluster}
Instead of workload simulation, resource utilization is collected from our production environment.
It has a range of 3 months, and each time tick is 5 minutes.
Compared with Alibaba or Microsoft, our cluster has only a relatively small scale, but our jobs are more stationary.
We adopt a series of m5 and g4 instances to support Spark and Deep Learning.
We build our system on an AWS-hosted v1.15.1 Kuberketes cluster,
which has a cluster-autoscaler component configured with auto-scaling groups of m5 series and g4d series instance.
Applications running on the system is a mix of different batch workloads such as deep learning, Spark, and HPC.

\subsection{Experimental Setup}

Generally the seasonality has day, week and year grain, but our workload trace only cross months, so we 
adopt a week period for our dataset and Philly's, and day period for Fuxi's. The test time range is represented in the
Figure \ref{training_datasets}.
We compare the result produced by the ARIMA, SES, LSTM, Transformer and static allocation, it shows that transformer outperforms other models.

\begin{table*}[t]
\caption{Training Datasets}
\label{training_datasets}
\vskip 0.15in
\begin{center}
\begin{small}
\begin{sc}
\begin{tabular}{cccc}
\toprule
Name & Time Range & Baseline Instance & Test Set Time Range\\
\midrule
  Our cluster trace & 2 months & g4.xlarge and m5.xlarge  & 1 week\\
  Microsoft Philly cluster trace & 3 months & g4.xlarge & 1 week\\
  Alibaba Fuxi cluster trace & 1 week & m5.24xlarge & 1 day \\
\bottomrule
\end{tabular}
\end{sc}
\end{small}
\end{center}
\vskip -0.1in
\end{table*}

\begin{figure*}[t]
  \centering
  \includegraphics[width=6.75in,height=4in]{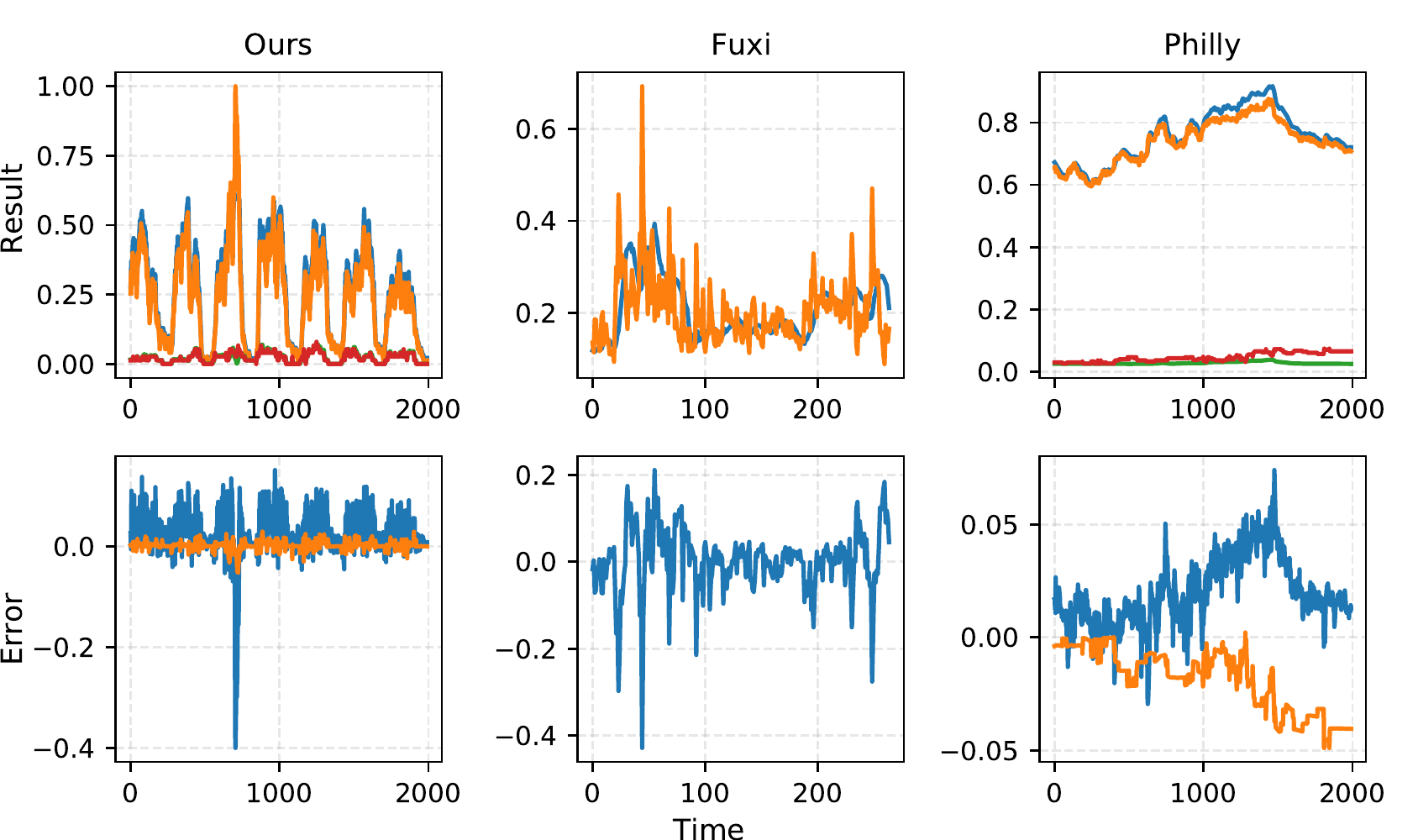}  
  \DeclareGraphicsExtensions.
  \caption{Predicted result and error curve. The orange lines stand for actual resource usage and blue lines stand for the predicted result in the result figures. In the error figures, the orange lines stand for rarely used instances error and blue lines stand for widely used instances error.}
  \label{exp_curve}
\end{figure*}

A ARIMA model is used to predict usage on the workloads trace. We selected the order of ARIMA
model using AIC and BIC to balance model complexity and
generalization. We used ARIMA(7, 1, 7) and a constant
trend to keep the model parsimonious. The fitted parameters
are then used on the full time series to make one-step ahead predictions.

We employed the automatic selection of the SES(Seasonal Exponential Smoothing) models to fit exponential models that had multiplicative components and 
evaluated possible  models prior to selecting the best-performing model to simulate the data.
The settings are $\alpha=0.5,\beta=0.001,\gamma=0.3,m=2016$ for ours and Philly's and $\alpha=0.5,\beta=0.005,\gamma=0.3,m=288$ for Fuxi's, where $\alpha$,$\beta$,$\gamma$ and $m$ stand for trend, level, seasonal level and period respectively.

The LSTM model has a stack of two LSTM layers
and a final linear layer to predict the instance count.
The LSTM layers encode sequential information
from input through the recurrent network. The fully-connected layer takes final output from the second LSTM layer
and outputs a vector for the instance counts.
The comparison in \cite{sak2014long} shows a network with two layers of LSTM can exceed state-of-the-art performance,
and after fine hyperparameter tuning, we find a hidden LSTM layer with a size of $51 \times 51$. Huber loss, Adam optimizer, and
a learning rate of 0.02 are used for training.

\subsection{Evaluation}

To measure the accuracy of the model, we use MSE(Mean Squared Error) as the evaluation metric. We extend the Mean Squared Error to PMSE(Positive Mean Squared Error),
where only predicted result larger than target will be computed. The equation follows:

\[ MSE(y,\hat{y}) = \frac{1}{n}\sum_{i=1}^{n}(y_{i}-\hat{y}_{i})^2 \]

\[ PMSE(y,\hat{y}) = \frac{1}{n}\sum_{i=1}^{n}(y_{i}-\hat{y}_{i})^2, \hat{y}_{i} > y_i \]

Our objective is to minimize the pending time of jobs and the number of instances for provisioning.
The instance launching time varies in accordance with different cloud providers.
The empirical initialization time is 5 minutes for AWS.
It may be longer if the the number of small instances is large,
because the AWS will split the larger instances into smaller instances
when there are no small instances available in a specific zone.
We don't scale down when predictor gives a underprovisioning scaling plan,
and adopt the PMSE our metric concerned with cost which means the overprovisioning part of our scaling plan.

\begin{figure*}[t]
  \centering
  \includegraphics[width=6.75in,height=2in]{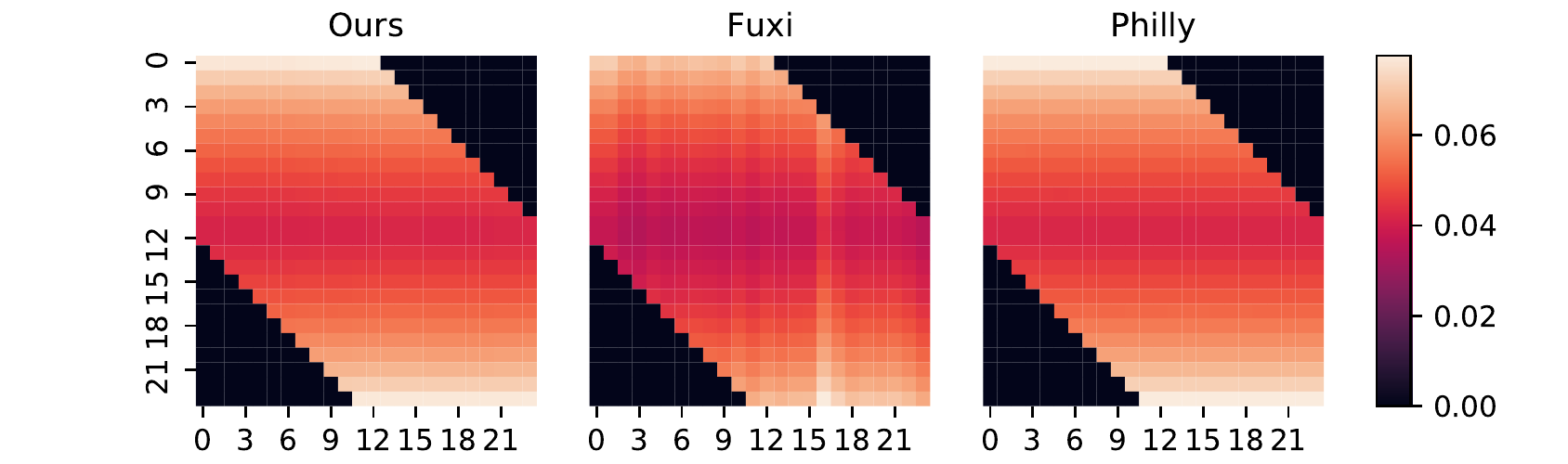}
  \DeclareGraphicsExtensions.
  \caption{Attention Heatmap of the first layer.
  The sliding widow contains 2 days of data points. Most of the attention is paid to the near data points. In the attention map of Fuxi, day at index 16 get most of the attention.}
  \label{attn_map}
\end{figure*}

Currently statistical models have no GPUs accelerating libraries. We test SES and ARIMA on CPUs only.
To adapt them to our multi-dimentional bucket counts,
we train each bucket count respectively and only list the training time for single bucket count.
It shows that statistical models are not competitive with GPU equipped neural networks
in speed on large dataset like Ours and Philly's. Statistical model work on single dimensional scalers.

After running the experiments, the predicted result and validation data are illustrated in Figure \ref{exp_curve}.
For clear seasonal data like ours and Fuxi, the weekly
workloads follow a consistent pattern. From the start,
the workloads gradually go up, after reaching the peak in the middle, the workloads go down.
At weekends, workloads keep at a relatively low level.
Daily workloads follow the pattern of working time.
The first peak value occurs at around 10 AM, the second peak occurs at about 3 PM.
There is no clear patterns for local workloads variance which fluctuates significantly.
Workloads of Philly does not have a clear period in the testing data but has a stable growing trend.
The experimental results are listed in Table \ref{exp_result}.
This experiment uses the MSE to reflect the accuracy of the regression model and use PMSE as the cost metric.

The results show that predictive model can save a large amount of cost compared to the maximum static allocation.
The Transformer model outperforms other models in the accuracy as well as cost saving.
The attention map is illustrated in Figure \ref{attn_map}.
This suggests that the attention is paid to the near time ticks, and the data point out of one day gains no attention.
Generally a stable workloads pattern as Philly is easier to estimate
while workloads that have unknown harsh peaks will make the model fail to give a reasonable result.
As shown in the Table \ref{exp_result}, the Transformer-based network achieves the best MSE and PMSE.
The ability to learn the complex pattern regardless the multiple seasonalities or local variance is also illustrated in Figure \ref{exp_curve}.
It does not cover the peak workload in the test of ours and Fuxi's, but perform well in the stable pattern of Philly's.
This peak load can be considered as an outlier, and the predictive scaler will downgrade to an on-demand scaler.

\begin{table}[t]
\caption{Experiments Results}
\label{exp_result}
\vskip 0.15in
\begin{center}
\begin{small}
\begin{sc}
\begin{tabular}{ccccc}
\toprule
Data & Model & MSE & PMSE & Time \\ 
\midrule
  \multirow{5}{*}{Ours} & Transformer & 2.51e-7 & 1.97e-7 & 44min\\
  & LSTM & 0.0047 & 0.0039 & 28min \\
  & SES & 0.0110 & 0.0046 & 37min \\
  & ARIMA & 0.027 & 0.0063 & 130min \\
  & Static & 0.7841 & 0.7841 \\
  \toprule
  \multirow{5}{*}{Fuxi}  & Transformer & 0.0021 & 0.0010 & 11min\\
  & LSTM & 0.0035 & 0.0074 & 8min \\
  & SES & 0.0026 & 0.0025 & 28s \\
  & ARIMA & 0.025 & 0.021 & 20min\\
  & Static & 0.8927 & 0.8927 \\
  \toprule
  \multirow{5}{*}{Philly} & Transformer & 0.0003 & 0.0004 & 38min\\
  & LSTM & 0.0040 & 0.0028 & 31min\\
  & SES & 0.1470 & 0.1470 & 46min \\
  & ARIMA & 0.020 & 0.093 & 84min \\
  & Static & 0.7054 & 0.7054 \\
\bottomrule
\end{tabular}
\end{sc}
\end{small}
\end{center}
\vskip -0.1in
\end{table}  
  
\section{Conclusion}

This paper proposes a method to address problems regarding 
predictive resources provisioning for elastic batch jobs.
Autoscaling of resources helps us to support customers' requirements
while keeping relative low cost.
All the resources requirements can be classified into resource buckets, which can simplify prediction model design.
The Transformer neural network can efficiently meet the needs of predicatively deciding scaling plans to satisfy drastically
varying job requirements on time. It has a good performance in learning the complex patterns of workload.
The attention mechanism can learn complex dependencies of various lengths from
time series data. The work presented demonstrates the feasibility of our approach in the context of ours,
Fuxi and Philly which validates the system is compatible with the Kubernetes and practical with modern cloud environment.

The neural network model is more expressive than other statistical methods.
Not only the numeric features but also other information can be embedded into
the model like the impact of external environments such as the user registrations, business events and holidays.
The additional features are likely to encode new information to the model.
They help the predictive model to build more confidence on peak workload.
In the future work we aim to provide a predictive model in accordance with the external information.

\bibliography{mlsys}
\bibliographystyle{mlsys2020}



\end{document}